\documentclass{article}

\usepackage{PRIMEarxiv}

\usepackage[utf8]{inputenc} 
\usepackage[T1]{fontenc}    
\usepackage{hyperref}       
\usepackage{url}            
\usepackage{booktabs}       
\usepackage{amsfonts}       
\usepackage{nicefrac}       
\usepackage{microtype}      
\usepackage{lipsum}
\usepackage{fancyhdr}       
\usepackage{graphicx}       
\graphicspath{{media/}}     

\usepackage{amsmath}
\usepackage{algpseudocode} 
\usepackage{algorithm}
\usepackage{booktabs}
\usepackage{multirow}
\usepackage{subcaption}
\usepackage{amsthm} 
\newtheorem{theorem}{Theorem}[section]

\newtheorem{lemma}[theorem]{Lemma}

\title{RAM: Replace Attention with MLP for Efficient Multivariate Time Series Forecasting
}

\author{
  Suhan Guo \\
  School of Artificial Intelligence \\
  Nanjing University \\
  Nanjing\\
  \texttt{shguo@smail.nju.edu.cn} \\
  \And
  Jiahong Deng \\
  School of Artificial Intelligence \\
  Nanjing University \\
  Nanjing\\
  \texttt{jiahongdeng@smail.nju.edu.cn} \\ 
  \And
  Yi Wei \\
  School of Intelligence Science and Technology \\
  Nanjing University \\
  Nanjing\\
  \texttt{ywei@smail.nju.edu.cn} \\ 
  \And
  Hui Dou \\
  Department of Computer Science and Technology \\
  Nanjing University \\
  Nanjing\\
  \texttt{huidou@smail.nju.edu.cn} \\ 
  \And
  Furao Shen* \\
  School of Artificial Intelligence \\
  Nanjing University \\
  Nanjing\\
  \texttt{frshen@nju.edu.cn} \\
  \And
  Jian Zhao* \\
  School of Electronic Science and Engineering \\
  Nanjing University \\
  Nanjing\\
  \texttt{jianzhao@nju.edu.cn} \\
}

\begin{document}
\maketitle

\begin{abstract}
Attention-based architectures have become ubiquitous in time series forecasting tasks, including spatio-temporal (STF) and long-term time series forecasting (LTSF). Yet, our understanding of the reasons for their effectiveness remains limited. In this work, we propose a novel pruning strategy, \textbf{R}eplace \textbf{A}ttention with \textbf{M}LP (RAM), that approximates the attention mechanism using only feedforward layers, residual connections, and layer normalization for temporal and/or spatial modeling in multivariate time series forecasting. Specifically, the Q, K, and V projections, the attention score calculation, the dot-product between the attention score and the V, and the final projection can be removed from the attention-based networks without significantly degrading the performance, so that the given network remains the top-tier compared to other SOTA methods. RAM achieves a $62.579\%$ reduction in FLOPs for spatio-temporal models with less than $2.5\%$ performance drop, and a $42.233\%$ FLOPs reduction for LTSF models with less than $2\%$ performance drop.
\end{abstract}


\section{Introduction}
A time series consists of data points recorded chronologically at fixed intervals. By leveraging the temporal trends within historical data, we can predict future data points, a task known as time series forecasting. This task finds wide applications in real-world scenarios such as electricity consumption forecasting \cite{sarmasTransferLearningStrategies2022}, tracking infectious disease spread \cite{dengColaGNNCrosslocationAttention2020}, predicting stock market fluctuations \cite{liStockMarketForecasting2020}, and forecasting traffic patterns \cite{wuConnectingDotsMultivariate2020}. 
Most real-world applications produce data that can be structured as a multivariate time series, often treated as a multi-channel signal. Depending on the domain, these channels are referred to as variables, nodes, or spatial dimensions. Studies diverge on how to handle inter-channel dependencies: some address channels independently, focusing on each channel as a distinct entity \cite{zhengTimeSeriesClassification2014, zengAreTransformersEffective2023, nieTimeSeriesWorth2023}, while others implicitly model correlations among sensors by projecting input signal vectors into an embedding space to integrate information across channels \cite{zhouInformerEfficientTransformer2021, wuAutoformerDecompositionTransformers2021, liuPyraformerLowcomplexityPyramidal2022}. 

Letting the lookback window and forecasting horizon be denoted as $\mathcal{T} $ and $\mathcal{H}$, respectively, we summarize mainstream multivariate time series forecasting problems into the following two types:
\begin{itemize}
    \item \textbf{Long-term Time Series Forecasting (LTSF)}: This task involves forecasting a sequence that significantly exceeds the length of the lookback window, represented as $\mathcal{T} \ll \mathcal{H}$. No explicit requirement is solicited for model structure. 
    \item \textbf{Spatio-Temporal Forecasting (STF)}: This task focuses on forecasting a sequence that is no longer than the lookback window, represented as $\mathcal{T} \geq \mathcal{H}$. The model structure should explicitly model temporal and spatial dependencies.
\end{itemize}
In both tasks, transformer-based architectures have become popular backbones for modeling temporal and/or spatial dependencies. The attention mechanism, a core component of transformers, is particularly effective at capturing associations across different time steps and channels, enabling the model to focus on relevant information in both dimensions adaptively.

Since the time and space complexity of the attention mechanism is quadratic in the input length, attention-based models for LTSF often aim to reduce the computational cost of attention scores from $\mathcal{O}(n^2)$ to $\mathcal{O}(n(\text{log}n)^2)$ or even $\mathcal{O}(n)$ \cite{kitaevReformerEfficientTransformer2020, wangLinformerSelfattentionLinear2020}. However, according to the Strong Exponential Time Hypothesis (SETH), the cost of calculating attention scores theoretically cannot be reduced below $\mathcal{O}(n^2)$ \cite{kelesComputationalComplexitySelfattention2023}. Furthermore, empirical studies suggest that simple linear transformations can outperform many transformer-based models in LTSF, raising questions about the attention mechanism’s effectiveness for temporal modeling \cite{zengAreTransformersEffective2023}. Recent research has shown that transformer-based models, when designed with reduced complexity, can still surpass linear transformations in LTSF \cite{nieTimeSeriesWorth2023, liuITransformerInvertedTransformers2024}. However, these models often lack cost-effectiveness, as the performance gains do not justify the increase in model parameters and computational demands.

Spatio-temporal data is a special type of multivariate time series that highlights both spatial and temporal dependencies. Measurements at each location are closely influenced by their historical values and those of nearby locations, making it essential to model both types of dependencies within the network explicitly \cite{atluriSpatiotemporalDataMining2018}. Many real-world applications can be represented as spatio-temporal data. A prominent example is traffic flow prediction, where the flow at a road intersection depends on the time of day and the traffic conditions at upstream and downstream intersections. Sensor proximities are often captured in a geographic adjacency map, assigning greater weights to edges between sensors that are closer in Euclidean distance \cite{liDiffusionConvolutionalRecurrent2018}.

Following the foundational works STGCN \cite{yuSpatiotemporalGraphConvolutional2018} and DCRNN \cite{liDiffusionConvolutionalRecurrent2018}, which formally defined the spatio-temporal forecasting task, researchers commonly address temporal and spatial dependencies separately. Temporal dependencies are typically modeled with CNNs \cite{wuGraphWaveNetDeep2019, wuConnectingDotsMultivariate2020}, LSTMs \cite{liDiffusionConvolutionalRecurrent2018, liDynamicSpatialtemporalGraph2021}, or attention mechanisms \cite{zhengGMANGraphMultiAttention2020, guoLearningDynamicsHeterogeneity2022}. Spatial dependencies, on the other hand, are framed as a graph learning problem. Because the geographic graph is insufficient to describe dependencies, a group of neural network models \cite{wuGraphWaveNetDeep2019, wuConnectingDotsMultivariate2020, songSpatialtemporalSynchronousGraph2020, yanLearningDynamicHierarchical2022} utilizes a graph mined from the data as the adjacency matrix for the downstream graph neural network module, referring to the graph as adaptive.

\begin{figure}
    \centering
    \includegraphics[width=0.5\linewidth]{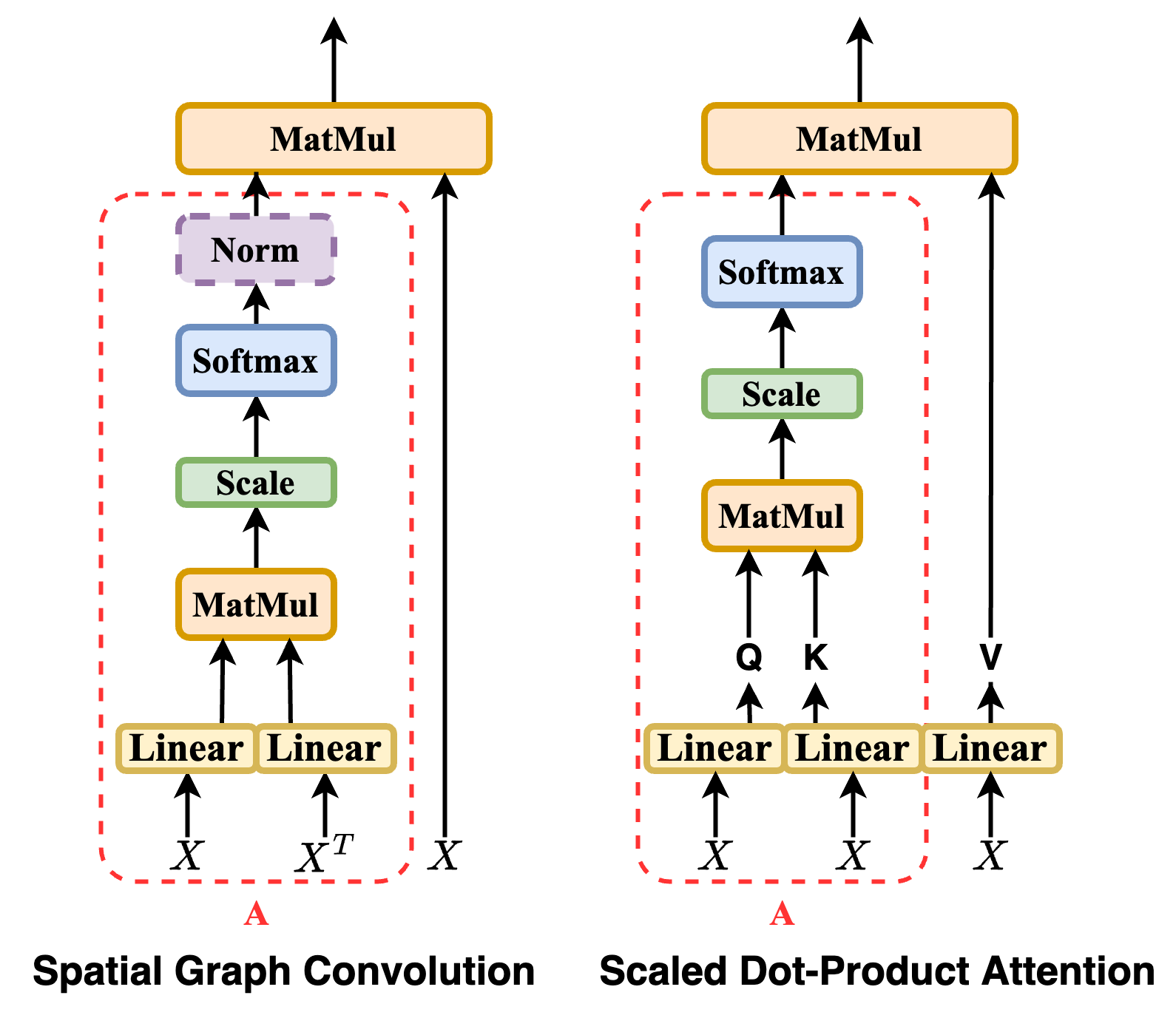}
    \caption{GCN as a modified attention mechanism in code implementation.}
    \label{fig:GCN=attention}
\end{figure}

To reduce computational costs, prior works primarily focus on pruning edges in either the adjacency matrices of spatial graphs or the weight matrices within temporal modules using sparsification algorithms \cite{duanLocalisedAdaptiveSpatialtemporal2023, chenUnifiedLotteryTicket2021, liSGCNGraphSparsifier2020, youEarlybirdGcnsGraphnetwork2022, zhengRobustGraphRepresentation2020}. Upon examining the code and formulas in popular models, we observe that the implementations of traditional attention layers and graph convolutional layers are closely related, as discussed in Section \ref{sec:model_spatial}. Consequently, we simplify the graph convolutional layers by reducing them to a modified attention mechanism, as illustrated in Figure \ref{fig:GCN=attention}.

To investigate both LTSF and TSF problems, we define models using original and/or modified attention mechanisms in their encoder structure to solve multivariate time series forecasting problems as \textbf{a}ttention-based \textbf{m}ultivariate \textbf{t}ime \textbf{s}eries \textbf{f}orecasting \textbf{m}odels (AMTSFM). In this work, we propose empirically proving the following statement: \textbf{ The attention mechanism for AMTSFM can be approximated using a Multi-Layer Perceptron (MLP) containing the feedforward, the residual connection, and the layer normalization.} Our investigation focuses on attention-based LTSF and STF models, as they represent the state-of-the-art (SOTA) in forecasting tasks, and our primary interest lies within the attention module. We identify two key motivations for our work. First, from a practical standpoint, pruning the attention module to an MLP can significantly reduce computational complexity and model storage, enabling deployment on resource-constrained devices and facilitating the model adaptation to datasets with many nodes. Second, from a research perspective, many SOTA models assume the attention module’s effectiveness without explicitly verifying its contribution. Ablation studies typically remove the entire module, but do not assess whether the Q, K, V projections and attention score matrices contribute as expected in multivariate time series forecasting. By ensuring that the accuracy drop is minimal, we seek to demonstrate that, aside from the embedding of input features, the core components of the AMTSFM are the MLP (containing residual connections, feedforward layers, and layer normalization) and that the scaled dot-product attention mechanism, which represents relationships between input time steps or nodes, is redundant.

One of the most closely related works is using adaptive graph sparsification to reduce the complexity of adaptive spatial-temporal graph neural networks \cite{duanLocalisedAdaptiveSpatialtemporal2023}. However, our focus differs in that we do not aim to develop an optimal sparsification algorithm for adjacency matrix in graph neural networks but rather to assess the effectiveness of the attention weight matrices in both temporal and spatial modules. By concentrating primarily on attention-based networks, we have found that these networks do not require spatial or temporal attention matrices during both the training and inference stages, as only a slight drop in performance is observed alongside a significant reduction in FLOPs. Our empirical findings suggest that LTSF and TSF models primarily rely on the embedding and linear transformation of input features, making explicitly modeling temporal and channel dependencies through the scaled dot-product attention mechanism less effective.



Our main contributions are summarised as follows:
\begin{itemize}
    \item To our knowledge, this is the first study to analyze the effectiveness of attention modules in AMTSFM. By replacing the attention mechanism in spatio-temporal models, such as STAEFormer and ASTGNN, with an MLP, we observe average increases in MAE, RMSE, and MAPE by $2.107\%$, $2.052\%$, and $3.065\%$, respectively, while reducing FLOPs and parameter counts by $62.247\%$ and $35.330\%$ on average. Similarly, for LTSF tasks on non-traffic datasets, we see average increases in MSE and MAE by $1.728\%$ and $0.742\%$ when using the PatchTST and iTransformer models, along with reductions in FLOPs and parameter counts of $52.173\%$ and $34.139\%$. These results underscore the potential efficiency gains from simplifying attention mechanisms.
    
    \item We propose an abstract structure for attention-based spatio-temporal networks by reducing the adaptive graph neural network to a modified attention mechanism. Through this abstraction, we observe that removing both spatial and temporal attention modules—components with the highest computational complexity—does not significantly degrade forecasting performance. Notably, the pruned model retains strong inference performance without requiring pre-training or fine-tuning.
    
    \item With further investigation, we discover that the attention in the decoder is the same as the attention in the encoder, which can be safely removed without degrading the performance. Replacing the attention mechanism with MLP without modifying the encoder structure will cause a performance drop of 
    $1.944\%, -0.254\%$, and $3.928\%$ in MAE, RMSE, and MAPE, and FLOPs drop $40.187\%$. However, removing the attention mechanism in both the encoder and decoder will cause a major performance drop. 

    \item Our empirical findings indicate that within the MLP structure used to approximate the encoder attention, the feedforward, and residual connection emerge as the core component of AMTSFM. This suggests that, for current multivariate datasets, the forecasting task can effectively be treated as a univariate forecasting problem, disregarding dependencies between time steps and nodes.
    
\end{itemize}

\section{Related Work}

\subsection{Attention-Based Spatio-Temporal Traffic Flow Forecasting}
Traffic flow forecasting is a critical component of intelligent transportation systems, requiring effective modeling of both spatial and temporal dependencies in traffic data. In recent years, GNN-based and Transformer-based methods have been widely explored to tackle this challenge. With the advent of GNNs, researchers have utilized the graph messaging mechanism to model spatial dependencies, improving forecasting accuracy by incorporating both spatial and temporal features. For example, GMAN \cite{zhengGMANGraphMultiAttention2020} employs spatial and temporal attention mechanisms to capture dynamic spatial patterns and complex temporal correlations in traffic data. Similarly, ASTGNN \cite{guoLearningDynamicsHeterogeneity2022} enhances this approach by introducing a multi-head self-attention-based temporal block alongside a dynamic graph convolution module to better capture traffic dynamics across time and space. Despite their effectiveness, GNN-based models have limitations in traffic prediction. For instance, the spatial dependencies between locations are highly dynamic, and distant locations may exhibit similar traffic patterns, which traditional GNN structures struggle to capture. To address this, PDFormer \cite{jiangPDFormerPropagationDelayaware2023} incorporates a delay-aware feature transformation module to handle time delays and uses dual graph masks to account for both short- and long-range spatial dependencies. Alternatively, rather than increasing model complexity, STAEformer \cite{liuSTAEformerSpatiotemporalAdaptive2023} leverages a spatio-temporal adaptive embedding applied to a vanilla Transformer, achieving SOTA performance with a simpler model structure. In this research, we investigate both GNN-based and Transformer-based models due to their close implementation similarities, aiming to examine their efficacy in traffic forecasting further.

\subsection{Transformer-Based Long-term Time Series Forecasting}
The Transformer architecture has become a leading framework for sequence-related tasks, including long-term time series forecasting. LogTrans \cite{liEnhancingLocalityBreaking2019} introduces convolutional self-attention layers within a LogSparse Transformer, capturing local context while reducing space complexity and addressing the memory limitations of standard Transformers. Informer \cite{zhouInformerEfficientTransformer2021} enhances predictive capacity and efficiency with a ProbSparse self-attention mechanism and a distillation operation to lower time and memory costs. Autoformer \cite{wuAutoformerDecompositionTransformers2021} presents a decomposition-based architecture using an Auto-Correlation mechanism to capture dependencies across series. Pyraformer \cite{liuPyraformerLowcomplexityPyramidal2022} uses a pyramidal attention module with inter-scale trees and intra-scale connections, achieving linear complexity. FEDformer \cite{zhouFEDformerFrequencyEnhanced2022} integrates Fourier-enhanced blocks following seasonal-trend decomposition to achieve linear complexity as well. PatchTST \cite{nieTimeSeriesWorth2023} leverages a channel-independent patch approach, segmenting the sequence into subseries-level patches that serve as input tokens, setting new benchmarks in forecasting accuracy. These studies primarily aim to optimize time and space complexity by introducing advanced attention mechanisms. In this paper, we question the necessity of attention mechanisms in long-term time series forecasting and examine the performance of Transformer-based models with and without these mechanisms.

\subsection{Effective of Attention Mechanism}
The Transformer architecture \cite{vaswaniAttentionAllYou2017} centers around the self-attention mechanism, which has a quadratic time and space complexity relative to the input length, enabling it to capture dependencies across sequence elements. Previous work \cite{kelesComputationalComplexitySelfattention2023} has shown that self-attention fundamentally requires quadratic time complexity under the Strong Exponential Time Hypothesis, though a linear-time approximation has been proposed, with complexity depending on an exponential polynomial order. Beyond computational cost, research has also examined self-attention’s effectiveness. Network rank, which indicates information flow across layers, is often used to evaluate model structure effectiveness \cite{fengRankDiminishingDeep2022}. Theoretical findings demonstrate that residual connections and MLPs help prevent the degradation of attention layers into rank-1 matrices, preserving the expressiveness of attention-based architectures \cite{dongAttentionNotAll2021}.

In addition to theoretical insights, practical studies have also scrutinized attention mechanisms. Zeng et al. \cite{zengAreTransformersEffective2023} question the efficacy of Transformer-based models in long-term time series forecasting, showing that a simple linear model, LTSF-Linear, surpasses complex Transformer models on multiple benchmarks. This finding suggests that the temporal modeling capabilities of Transformers in time series may be overestimated. Duan et al. \cite{duanLocalisedAdaptiveSpatialtemporal2023} propose a method to localize Adaptive Spatial-Temporal Graph Neural Networks by sparsifying their spatial graph adjacency matrices, demonstrating that while spatial dependencies are vital during training, they can be largely ignored during inference. In this paper, we examine both spatial-temporal traffic flow forecasting and long-term time series forecasting, investigating whether attention matrices in the attention module are necessary for both training and inference.

\section{Methods}
\subsection{Definition of Multivariate Time Series Data Structure}
The multivariate time series is defined as a sequence of discrete signal matrices $\mathcal{X} = \{X_{t-T+1}, \dots, X_{t}\}$. Given the lookback window $T=\mathcal{T}$, the historical signals are represented as $\mathcal{X} = \{X_{t-\mathcal{T} + 1}, \dots, X_{t}\} \in \mathbb{R}^{\mathcal{T} \times N \times C}$, where $X_t \in \mathbb{R}^{N \times C}$ represents a signal matrix at time $t$ for $N$ nodes with $C$ features per node. For Spatio-Temporal Forecasting (STF), the input data also includes a network structure, abstracted as a graph $\mathcal{G} = (\mathcal{V}, \mathcal{E})$. This graph $\mathcal{G}$ represents a network containing $|\mathcal{V}|=N$ nodes (e.g., traffic detectors, observation stations) and $|\mathcal{E}|=M$ edges, which describe the connections between node pairs, either indicating presence or weighted strengths of connections. Given the forecasting window as $\mathcal{H}$ and the learnable parameters as $\theta$, the goal is to learn the mapping $\mathcal{F}(\cdot;\theta)$ that predicts the future sequence, as defined by:
\begin{align}
    \{X_{t+1}, \dots, X_{t+\mathcal{H}}\} = \mathcal{F}(\mathcal{X}; \theta, \mathcal{G}).
\end{align}
For LTSF, the graph $\mathcal{G}$ has the following property:
\begin{align}
    \mathcal{E} \sim \text{Uniform}(1, N).
\end{align}
Since no prior information on variate dependencies is available, the graph input is not used in LTSF.

\subsection{Softmax Degeneration}
The multi-head attention mechanism captures correlations between input elements. When these associations are trivial, the attention mechanism may degenerate into a simple identity mapping, as demonstrated in Lemma \ref{lemma:att_eq_identity} inspired by \cite{velickovicSoftmaxNotEnough2024}. With skip connections incorporated, the degenerated attention projection simplifies to $y = 2x$. Consequently, removing the attention mechanism does not significantly impact the network, provided that downstream parameters can adjust to scale the input by a factor of 2.

\begin{lemma} \label{lemma:att_eq_identity}
    \textbf{Softmax is invariant to uniform distribution input.}
    Let $\beta \geq 0$ be the temperature and $\{S_1, \dots, S_T\} \in \mathbb{R}^{T}$ be the collection of $T$ logits as a parameter and input to the softmax function. Bound the input logits with some $L, U \in \mathbb{R}$, we have $L \leq S_i \leq U, i = 1, \dots, T$. If the input exponent follows the uniform distribution $S_i \sim \text{Uniform}(k, kT)$ with $k \in \mathbb{R}$ and $k \neq 0$, the output of the softmax also follows the uniform distribution as $M_i \sim \text{Uniform}(1, T)$.

    \begin{proof}
        Define the attention output assigned to input $i$ as $M_i = \text{Softmax}(\{e^{S_1}, \dots, e^{S_T}\}; \beta) \in [0, 1]$. Then, it has bound as:
        \begin{align*}
             \frac{\exp(\beta L)}{T \exp(\beta U)} \leq   M_i = \frac{\exp(\beta S_i)}{\sum_{j=1}^{T} \exp(\beta S_j)} \leq \frac{\exp(\beta U)}{T \exp(\beta L)},
        \end{align*}
        because the exponential function is monotonically increasing.
        Let $\delta = U - L$, we have:
        \begin{align*}
            \frac{1}{T} \exp(-\beta \delta) \leq M_i \leq \frac{1}{T} \exp(\beta \delta).
        \end{align*}
        Since $S_i$ follows a uniform distribution, then $\delta = U - L = 0$,
        hence $\exp(-\beta \delta) = \exp(\beta \delta) = 1$, we have:
        \begin{align*}
            \frac{1}{T} \leq M_i \leq \frac{1}{T},
        \end{align*}
        showing that $M_i = \frac{1}{T}$, which is the PDF of $\text{Uniform}(1, T)$.
    \end{proof}
    
\end{lemma}

\subsection{Modeling the Temporal Association}
Attention is a core component in many spatio-temporal models, where it captures associations between adjacent and non-adjacent positions in the sequence, assigning weights to these relationships and producing a weighted sum as output. 
It has also been adapted in Long-Term Time Series Forecasting (LTSF) models to model temporal dependencies efficiently. 
Multi-head attention consists of several parallel layers, each attending to information from different representation subspaces at various positions. For each attention head, the self-attention score $S \in \mathbb{R}^{\mathcal{T} \times \mathcal{T}}$ is calculated by performing a dot product between the query and key matrices, $Q, K \in \mathbb{R}^{\mathcal{T} \times d_{\text{head}}}$, along the temporal dimension, as shown below:
\begin{align} \label{eq:attention_score}
    Q &= E_n^{(l-1)}W^Q, \quad K = E_n^{(l-1)}W^K, \nonumber\\
    S &= \text{softmax}(\frac{QK^T}{\sqrt{d_{\text{head}}}}).
\end{align}
The $E_n^{(l-1)} \in \mathbb{R}^{\mathcal{T} \times d_{\text{head}}} $ represents the input for the current module, and the weighted sum is obtained using:
\begin{align}
    H = S(E_n^{(l-1)}W^V) = SV, \label{eq:attention_value}
\end{align}
where $V \in \mathbb{R}^{\mathcal{T} \times d_{\text{head}}}$ is the value. The weighted sums from $K$ heads are concatenated, denoted as $[\cdot]$, and projected with $W^O \in \mathbb{R}^{d_{\text{model}} \times d_{\text{model}}}$, shown as:
\begin{align}
    E_t^{(l)} = \left[ H_1, \dots, H_K \right] W^O. \label{eq:attention_output}
\end{align}
Summarizing the above equations, we recapitulate the temporal module for attention-based models as follows:
\begin{align}
    E_n^{(l)} = \text{TemporalAttention}(E_n^{(l-1)}).
\end{align}

\subsection{Modeling the Spatial Association} \label{sec:model_spatial}
The geographical adjacency matrix, $\mathcal{A}_s \in \mathbb{R}^{N \times N}$, and the dynamic adjacency matrix, $\mathcal{A}_d \in \mathbb{R}^{N \times N}$, are commonly used in spatio-temporal research. In traffic datasets, the geographical adjacency matrix is often constructed by transforming distances between detectors into connection weights. 
However, this matrix is not always available for every dataset and may not fully capture node relationships. To address these limitations, the dynamic adjacency matrix is introduced. This matrix is learned from the input node feature $Z_t^{(l-1)} \in \mathbb{R}^{N \times d_{\text{model}}}$, enabling a more adaptable representation of node dependencies, i,e.,
\begin{align}
    \mathcal{A}_d = \text{softmax}\left(\frac{(Z_t^{(l-1)} W^Q)(Z_t^{(l-1)} W^K)^T}{\sqrt{d_{\text{model}}}}\right). \label{eq:get_dynamic_adj}
\end{align}
The $W^Q, W^K$, and $W^V$ represent the learnable weights in linear projection layers. With an adjacency matrix, a Graph Convolutional Network (GCN) layer is defined as:
\begin{align}
    \tilde{\mathcal{A}} &= D^{-\frac{1}{2}}\mathcal{A}D^{-\frac{1}{2}}, \label{eq:gcn_norm}\\
    Z_t^{(l)} &= \sigma({\tilde{\mathcal{A}}} Z_t^{(l-1)} \Theta), \label{eq:perform_gcn}
\end{align}
where $\mathcal{A}$ can be either a geographical or dynamic adjacency matrix and $\sigma(\cdot)$ represents the activation function.

In practice, the GCN module is frequently simplified to a modified single-head attention mechanism applied along the node dimension. When dynamically generating the adjacency matrix, we observe that Eq. \eqref{eq:get_dynamic_adj} is analogous to the attention score generation in Eq. \eqref{eq:attention_score}. For message propagation, the typical matrix normalization step, shown in Eq. \eqref{eq:gcn_norm}, is often omitted. Consequently, Eq. \eqref{eq:perform_gcn} aligns with Eq. \eqref{eq:attention_value} and Eq. \eqref{eq:attention_output}, where the value projection layer is removed and an activation function is added.

In LTSF tasks, if the channel-mixing assumption is applied, researchers model spatial associations within the embedding layers without explicitly using a GCN, as illustrated above. When channel independence is assumed, variables are treated as independent. However, certain studies, like the iTransformer \cite{liuITransformerInvertedTransformers2024}, emphasize spatial dependencies. iTransformer introduces an inverted attention layer, which resembles our spatial attention structure, while temporal associations are managed using feed-forward layers.

In summary, the spatial module for attention-based models is: 
\begin{align}
   Z_t^{(l)} = \text{SpatialAttention}(Z_t^{(l-1)})
\end{align}

\subsection{Abstract Structure of AMTSFM}
Since both temporal and spatial associations are modeled using the attention mechanism in implementation, the spatio-temporal networks can be reduced to three components, including the embedding module, the encoder, and the decoder. 

\subsubsection{Embedding}
The naive embedding layer is a linear projection that maps raw input data to a target dimension. In some cases, additional temporal patterns, such as day-of-week or time-of-day, are embedded to enhance the model’s temporal awareness. Others further improve spatial awareness by embedding the spatial dimension order into a higher dimension and adding it to the input features to strengthen the model’s understanding of spatial connections. 
In general, the embedding layer can be summarized as:
\begin{align}
    E_0 &= \textit{Embedding}(\mathcal{X}),
\end{align}
where input is mapped to feature $E_0 \in \mathbb{R}^{\mathcal{T} \times N \times d_{\text{model}}}$.

\subsubsection{Encoder}
The encoder is composed of modules that alternately model spatial and temporal associations. To handle the multiplications between tensors efficiently, which are processed as batched matrix multiplications, we introduce two helper functions: a split function $\ominus(\cdot; D)$ and a stack function $\oplus(\cdot; D)$, where $D$ represents the specified dimension along which the operations occur. These functions are defined as follows:
\begin{align}
    \{ E_{n=1}^{(0)}, \dots, E_{n=N}^{(0)} \} &= \ominus(E_0; N), \\
    E_0 &= \oplus(E_{n=1}^{(0)}, \dots, E_{n=N}^{(0)}; N)
\end{align}
where temporal input $E_{n}^0 \in \mathbb{R}^{\mathcal{T} \times d_{\text{model}}}$ is obtained by spliting the input feature $E_0$ at the node dimension $N$. The $E_0 \in \mathbb{R}^{\mathcal{T} \times N \times d_{\text{model}}}$ represents the temporal outputs stacked at the node dimension.

To distinguish between TSF and LTSF, we define an indicator function as:
\begin{align}
    \mathbb{I}(\tau = 1),
\end{align}
where $\tau = 1$ and $\tau = 2$ represent TSF and LTSF, respectively. The modeling of spatial dependencies is default for TSF but optional for LTSF. Defining the final output from the temporal module to be the input to the spatial module, denoted as $Z_0 = E_L$, the encoder is defined as: 
\begin{align}
    E_L &= \mathcal{O}_{l=1}^{L} \oplus(\text{TemporalAttention}_l(\ominus(E_0; N)); N ), \\
    Z_L &= \mathbb{I}(\tau = 1)\mathcal{O}_{l=1}^{L} \oplus(\text{SpatialAttention}_l(\ominus(Z_0; T)); T), 
\end{align}
where the $\mathcal{O}_{l=1}^{L}$ represents a composition of functions over $L$ layers. Note that each attention module is interleaved with layer normalization, feedforward projection, dropout, and residual connections. For simplicity, we omit these elements in the equation above, but they are detailed in Section \ref{sec:MLP_replace_attention}.

\subsubsection{Decoder}
Some studies retain the original form for decoders, using a masked version of the attention module to ensure predictions depend solely on previously known outputs, denoted as $\delta = 1$. In other models, the decoder is simplified to a linear projection or convolutional layer to mitigate overfitting, denoted as $\delta = 2$, with $\text{TemporalProj}(\cdot)$ and $\text{SpatialProj}(\cdot)$ represents these projection layers. We summarize the decoder into the following equation:
\begin{align}
    E_{K} &= \mathbb{I}(\delta = 2)\oplus(\text{TemporalProj}(\ominus(E_0; N)); N) + \mathbb{I}(\delta = 1)\mathcal{O}_{l=1}^{K} \oplus(\text{TemporalAttention}_l(\ominus(E_0; N)); N ), \\
    Z_{K} &= \mathbb{I}(\delta = 2)\oplus(\text{SpatialProj}_l(\ominus(Z_0; T);N) + \mathbb{I}(\delta = 1)\mathcal{O}_{l=1}^{K} \oplus(\text{SpatialAttention}_l(\ominus(Z_0; T)); T).
\end{align}


\begin{figure}
    \centering
    \includegraphics[width=\linewidth]{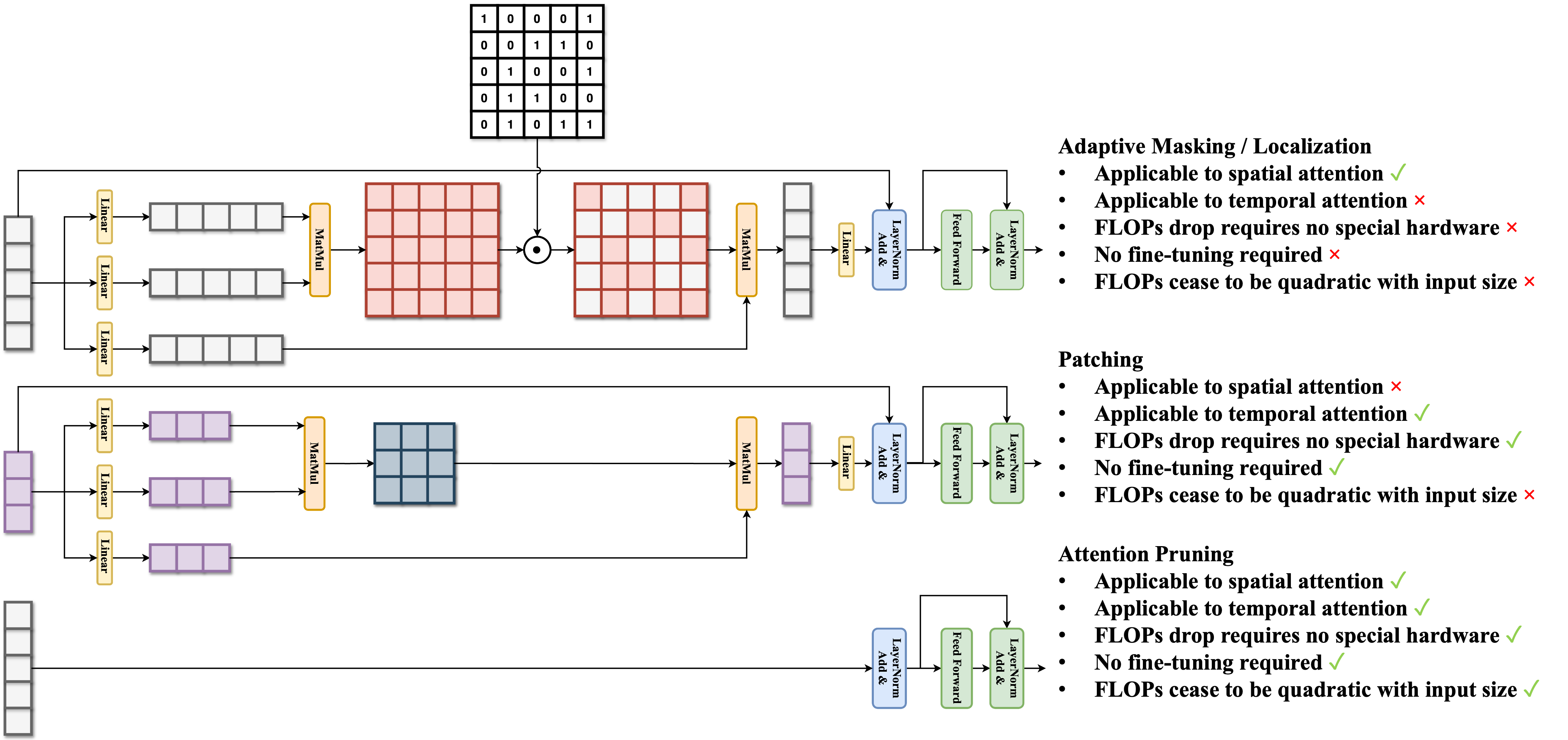}
    \caption{Comparison between Adaptive Making (Localization), Patching, and Attention Pruning strategies. For attention pruning, we remove the attention layer from the attention block, transforming it into an MLP block with feedforward, residual connection, and layer normalization layers.}
    \label{fig:mlp_replace_attention}
\end{figure}

\subsection{Replace Attention with MLP} \label{sec:MLP_replace_attention}
We remove the $\text{TemporalAttention}(\cdot)$ and $\text{SpatialAttention}(\cdot)$ modules while retaining the remaining operations in each attention block, as shown in Figure \ref{fig:mlp_replace_attention}. We discover that the drop in performance is moderate, while the decrease in FLOPs is considerable. Besides the core attention mechanism, there are auxiliary operations, including feedforward, layer normalization, dropout, and residual connections. Taking the temporal module input $E_n^{(l-1)}$ as an example, these operations are organized as follows:
\begin{align}
    R_n^{(l)} &= \text{LayerNorm}(\text{Dropout}(E_n^{(l-1)})) + E_n^{(l-1)}), \label{eq:ln1} \\
    F_n^{(l)} &= \sigma(R_n^{(l)}W^F)W^B, \label{eq:ffn} \\ 
    E_n^{(l)} &= \text{LayerNorm}(\text{Dropout}(F_n^{(l)}) + R_n^{(l)}).\label{eq:ln2} 
\end{align}
$W^F \in \mathcal{R}^{d_{\text{model}} \times d_{\text{feedforward}}}$ and $W^B \in \mathcal{R}^{d_{\text{feedforward}} \times d_{\text{model}}}$ are two linear projection weights for the feedforward layers, which maps the current feature to higher dimension and back. We summarize the Eq. \eqref{eq:ln1}, Eq. \eqref{eq:ffn} and Eq. \eqref{eq:ln2} as:
\begin{align}
    E_n^{(l)} &= \text{ResNormFFN}(E_n^{(l-1)}), \\
    Z_t^{(l)} &= \text{Res}(Z_t^{(l-1)}),
\end{align}
for temporal and spatial modules, respectively. Then, the encoder can be reformulated as:
\begin{align}
    E_L &= \mathcal{O}_{l=1}^{L} \oplus(\text{ResNormFFN}_l(\ominus(E_0; N)); N ), \\
    Z_L &= \mathbb{I}(\tau = 1)\mathcal{O}_{l=1}^{L} \oplus(\text{ResNormFFN}_l(\ominus(Z_0; T)); T). 
\end{align}
Notice that the original computational cost of the attention mechanism is $\mathcal{O}(n^2)$, where $n$ represents the sequence length or the number of nodes. This quadratic complexity arises because the self-attention mechanism computes pairwise interactions between all elements in the sequence. However, considering that the dimension of feature maps is much smaller than the number of nodes or sequence length, the quadratic dependency can be a significant bottleneck in large-scale problems. In contrast, the MLP-replace-attention version has a computational cost of $\mathcal{O}(d_{\text{model}})$, where $d_{\text{model}}$ is the model’s feature dimension. By eliminating the quadratic dependency on the sequence length or the number of nodes, this modification substantially reduces the overall computational complexity. This change results in a significant decrease in FLOPs while maintaining comparable forecasting performance.

\section{Experiments} 
We conduct experiments to show that 1) spatial and temporal attention mechanisms can be removed without significant performance loss for STF, and 2) attention modules are not crucial for modeling temporal dependencies in LTSF.

\subsection{Baseline models}
We selected two recent attention-based spatio-temporal network models, ASTGNN \cite{guoLearningDynamicsHeterogeneity2022} and STAEFormer \cite{liuSTAEformerSpatiotemporalAdaptive2023}, as well as two LTSF models, PatchTST \cite{nieTimeSeriesWorth2023} and iTransformer \cite{liuITransformerInvertedTransformers2024}. Their key features are:
\begin{itemize}
    \item \textbf{ASTGNN}: Models both traffic data periodicity and spatial heterogeneity using global/local periodic tensors and spatial positional embeddings, capturing spatial and temporal correlations via self-attention.
    \item \textbf{STAEformer}: Introduces spatio-temporal adaptive embedding to capture complex spatio-temporal relations alongside temporal periodicity and feature embeddings. 
    \item \textbf{PatchTST}: Reduces multivariate time series to single variate through channel independence and segments input into subseries-level patches to lower computational cost. 
    \item \textbf{iTransformer}: Applies attention to the spatial dimension and feedforward networks in the temporal dimension to capture multivariate correlations and nonlinearity.
\end{itemize}

\subsection{Datasets}
The MLP-replace-attention models are evaluated on real-world spatio-temporal datasets, including METR-LA, PEMS-BAY, PEMS04, PEMS07, and PEMS08, where DCRNN \cite{liDiffusionConvolutionalRecurrent2018} introduced the first two, and STSGCN \cite{songSpatialtemporalSynchronousGraph2020} proposed the rest. For long-term time series forecasting, we evaluate using 8 popular multivariate datasets \cite{nieTimeSeriesWorth2023}. Detailed statistics are provided in Table \ref{tab:datasets_stats}.

\begin{table}[h]
\centering
\caption{Dataset Statistics}
\label{tab:datasets_stats}
\resizebox{1\textwidth}{!}{
\begin{tabular}{c|cccccccc}
\toprule
\multicolumn{9}{c}{STF} \\
\midrule
Datasets & METR-LA & PEMS-BAY & PEMS04 & PEMS07 & PEMS08 &  &  &  \\
\midrule
Nodes & 207 & 325 & 307 & 883 & 170 &  &  &  \\
Time Steps & 34,272 & 52,116 & 16,992 & 28,224 & 17,856 &  &  &  \\
Time Range & 03/2012-06/2012 & 01/2017-05/2017 & 01/2018-02/2018 & 05/2017-08/2017 & 07/2016-08/2016 &  &  &  \\
\midrule
\multicolumn{9}{c}{LTSF} \\
\midrule
Datasets & Weather & Traffic & Electricity & ILI & ETTh1 & ETTh2 & ETTm1 & ETTm2 \\
\midrule
Nodes & 21 & 862 & 321 & 7 & 7 & 7 & 7 & 7 \\
Time Steps & 52,696 & 17,544 & 26,304 & 966 & 17,420 & 17,420 & 69,680 & 69,680 \\
\bottomrule
\end{tabular}}
\end{table}

For STF, the METR-LA and PEMS-BAY datasets are split into 70\% for training, 10\% for validation, and 20\% for testing. We use the past 1 hour (12 steps) of data to predict traffic flow for the next 15 minutes (3 steps), 30 minutes (6 steps), and 1 hour (12 steps). Consistent with previous works, we report point estimates of metrics for the three prediction lengths (steps 3, 6, and 12), along with the ``average'' estimate over the 1-hour forecast. For PEMS04, PEMS07, and PEMS08, the data is split 60\% for training, 20\% for validation, and 20\% for testing, using the past 12 steps to predict 12 steps forward. Averaged metrics are reported. To focus on the attention mechanism, networks are trained using their original data normalization and embedding versions.

We follow the standard protocol from previous works for dataset splitting, dividing all datasets into training, validation, and test sets chronologically for LSTF. For the ETT datasets, we use a $6:2:2$ split ratio, while for other datasets, a $7:1:2$ ratio is adopted. For the illness dataset, the lookback window is set to 36, and the forecasting horizons are 24, 36, 48, and 60. For the other datasets, the lookback window is 96, with forecasting horizons of 96, 192, 336, and 720.

\subsection{Setting and metrics}
Experiments are conducted on a machine with eight NVIDIA GeForce RTX 2080 GPUs using Ubuntu 18.04. We adhere to the loss function, Pytorch, and Python versions from the original papers. For STF datasets, evaluation metrics include mean absolute error (MAE), root mean squared error (RMSE), and mean absolute percentage error (MAPE), which are reported on the re-transformed original scale. MAPE results are reported as percentages. Missing values are excluded from training and inference. For LTSF datasets, MAE and RMSE are reported as in the original papers. Computational cost is measured in FLOPS (floating-point operations per second). All experiments are repeated five times, and average results are reported.

\subsection{Results}
We summarize our major experiments using spatio-temporal data in Section \ref{sec:results_STF} and multivariate data in Section \ref{sec:results_LTSF}. To validate our main argument, we first train the selected networks on their corresponding datasets to obtain baseline metrics. Then, we replace all attention modules in the encoder structure with an MLP and re-run the experiments with the same settings. For MAE, RMSE, and MAPE metrics, we include the standard deviation from five runs. The FLOPS ($\downarrow$) metric compares the inference FLOPS of the bias-replace-attention model with the original, showing the percentage decrease. The Params ($\downarrow$) metric indicates the parameter difference between the original model and the MLP-replace-attention model.

\subsubsection{Spatio-Temporal Forecasting} \label{sec:results_STF}

\begin{table}[h]
\centering
\caption{Performance of STAEformer on traffic datasets. Columns represent (1) the original model, (2) the MLP-replace-attention model, and (3) the performance improvement between them. Improvements in MAE, RMSE, and MAPE are highlighted in bold.}
\label{tab:staeformer_traffic}
\resizebox{1\textwidth}{!}{
\begin{tabular}{cc|ccc|ccc|ccccc}
\toprule
\multicolumn{2}{c|}{STAEformer} & \multicolumn{3}{c|}{Original Model} & \multicolumn{3}{c|}{MLP-replace-attention} & \multicolumn{5}{c}{Improvement} \\ \midrule
 &  & MAE & RMSE & MAPE(\%) & MAE & RMSE & MAPE(\%) & MAE($\uparrow$) & RMSE($\uparrow$) & MAPE($\uparrow$) & FLOPS($\downarrow$) & Params($\downarrow$) \\ 
 \midrule
PEMS04 & Avg. & $18.229\pm0.032$ & $30.239\pm0.114$ & $12.047\pm0.038$ & $18.467\pm0.015$ & $30.19\pm0.027$ & $12.115\pm0.025$ & 1.307\% & 0.162\% & 0.566\% & 64.067\% & 41.193\% \\
\midrule
PEMS07 & Avg. & $19.234\pm0.067$ & $32.75\pm0.138$ & $8.061\pm0.037$ & $19.718\pm0.044$ & $33.463\pm0.128$ & $8.224\pm0.033$ & 2.517\% & 2.176\% & 2.022\% & 74.299\% & 29.252\% \\
\midrule
PEMS08 & Avg. & $13.479\pm0.015$ & $23.276\pm0.068$ & $8.854\pm0.017$ & $13.686\pm0.041$ & $23.365\pm0.131$ & $9.052\pm0.034$ & 1.536\% & 0.385\% & 2.242\% & 60.309\% & 45.622\% \\
\midrule
\multirow{4}{*}{METR-LA} & 15min & $2.656\pm0.013$ & $5.119\pm0.051$ & $6.898\pm0.045$ & $2.823\pm0.004$ & $5.58\pm0.014$ & $7.718\pm0.032$ & 6.282\% & 8.998\% & 11.883\% & \multirow{4}{*}{61.399\%} & \multirow{4}{*}{44.334\%} \\
 & 30min & $2.967\pm0.018$ & $6.023\pm0.06$ & $8.153\pm0.078$ & $3.184\pm0.003$ & $6.631\pm0.015$ & $9.361\pm0.04$ & 7.293\% & 10.097\% & 14.825\% & & \\
 & 60min & $3.34\pm0.014$ & $7.023\pm0.047$ & $9.741\pm0.092$ & $3.546\pm0.002$ & $7.595\pm0.009$ & $10.943\pm0.043$ & 6.160\% & 8.146\% & 12.339\% & & \\
 & Avg. & $2.939\pm0.014$ & $5.989\pm0.05$ & $8.08\pm0.067$ & $3.126\pm0.002$ & $6.526\pm0.013$ & $9.129\pm0.036$ & 6.379\% & 8.961\% & 12.980\% & & \\
 \midrule
\multirow{4}{*}{PEMS-BAY} & 15min & $1.302\pm0.003$ & $2.774\pm0.005$ & $2.748\pm0.016$ & $1.356\pm0.002$ & $2.884\pm0.014$ & $2.884\pm0.009$ & 4.134\% & 3.971\% & 4.931\% & \multirow{4}{*}{64.509\%} & \multirow{4}{*}{40.675\%} \\
 & 30min & $1.603\pm0.005 $& $3.667\pm0.011$ & $3.606\pm0.038$ & $1.678\pm0.002$ & $3.811\pm0.009$ & $3.819\pm0.008$ & 4.664\% & 3.935\% & 5.915\% & & \\
 & 60min & $1.865\pm0.011$ & $4.312\pm0.019$ & $4.376\pm0.073$ & $1.95\pm0.002$ & $4.476\pm0.005$ & $4.591\pm0.004$ & 4.548\% & 3.813\% & 4.929\% & & \\
 & Avg. & $1.539\pm0.004$ & $3.55\pm0.014$ & $3.449\pm0.034$ & $1.608\pm0.002$ & $3.693\pm0.008$ & $3.637\pm0.005$ & 4.482\% & 4.034\% & 5.437\% & & \\ \bottomrule
\end{tabular}}
\end{table}

\begin{table}[h]
\centering
\caption{Performance of ASTGNN on traffic datasets. Columns represent (1) the original model, (2) the MLP-replace-attention model, and (3) the improvement between them. MAE, RMSE, and MAPE improvements are highlighted in bold.}
\label{tab:astgnn_traffic}
\resizebox{1\textwidth}{!}{
\begin{tabular}{cc|ccc|ccc|ccccc}
\toprule
\multicolumn{2}{c|}{ASTGNN} & \multicolumn{3}{c|}{Original Model} & \multicolumn{3}{c|}{MLP-replace-attention} & \multicolumn{5}{c}{Improvement} \\ \midrule
 &  & MAE & RMSE & MAPE(\%) & MAE & RMSE & MAPE(\%) & MAE($\uparrow$) & RMSE($\uparrow$) & MAPE($\uparrow$) & FLOPS($\downarrow$) & Params($\downarrow$)\\ 
 \midrule
PEMS04 & Avg. & $18.492\pm0.055$ & $31.168\pm0.226$ & $12.406\pm0.043$ & $18.49\pm0.038$ & $31.002\pm0.196$ & $12.428\pm0.084$ & $\textbf{-0.011\%}$ & -0.533\% & 0.177\% & 53.542\% & 31.492\% \\
\midrule
PEMS07 & Avg. & $18.34\pm0.089$ & $32.04\pm0.167$ & $7.668\pm0.044$ & $18.714\pm0.068$ & $32.928\pm0.124$ & $7.824\pm0.027$ & 2.039\% & 2.772\% & 2.034\% & 66.133\% & 25.465\% \\
\midrule
PEMS08 & Avg. & $12.79\pm0.07$ & $22.634\pm0.079$ & $8.712\pm0.082$ & $12.958\pm0.073$ & $22.972\pm0.14$ & $8.896\pm0.212 $& 1.314\% & 1.493\% & 2.112\% & 57.869\% & 32.128\% \\
\midrule
\multirow{4}{*}{METR-LA} & 15min & $3.306\pm0.015$ & $9.164\pm0.038$ & $7.508\pm0.07$ & $3.326\pm0.008$ & $9.128\pm0.024$ & $7.626\pm0.073$ & 0.605\% & \textbf{-0.393\%} & 1.572\% & \multirow{4}{*}{58.795\%} & \multirow{4}{*}{31.803\%} \\
 & 30min & $4.308\pm0.017$ & $11.656\pm0.031$ & $9.572\pm0.14$ & $4.328\pm0.017$ & $11.614\pm0.033$ & $9.668\pm0.075$ & 0.464\% & \textbf{-0.360\%} & 1.003\% & & \\
 & 60min & $5.626\pm0.022$ & $14.39\pm0.055$ & $12.152\pm0.206$ & $5.698\pm0.035$ & $14.436\pm0.037$ & $12.254\pm0.12$ & 1.280\% & 0.320\% & 0.839\% & & \\
 & Avg. & $4.264\pm0.014$ & $11.624\pm0.029$ & $9.444\pm0.13$ & $4.302\pm0.02$ & $11.63\pm0.021$ & $9.55\pm0.075$ & 0.891\% & 0.052\% & 1.122\% & & \\
 \midrule
\multirow{4}{*}{PEMS-BAY} & 15min & $1.298\pm0.004$ & $2.844\pm0.008$ & $2.752\pm0.027$ & $1.3\pm0.006$ & $2.84\pm0.017$ & $2.726\pm0.021$ & 0.154\% & \textbf{-0.141\%} & \textbf{-0.945\%} & \multirow{4}{*}{61.151\%} & \multirow{4}{*}{31.339\%} \\
 & 30min & $1.61\pm0.009$ & $3.816\pm0.022$ & $3.718\pm0.044$ & $1.612\pm0.007$ & $3.79\pm0.028$ & $3.674\pm0.042$ & 0.124\% & \textbf{-0.681\%} & \textbf{-1.183\%} & & \\
 & 60min & $1.882\pm0.01$ & $4.548\pm0.038$ & $4.586\pm0.057$ & $1.876\pm0.01$ & $4.502\pm0.037$ & $4.486\pm0.051$ & \textbf{-0.319\%} & \textbf{-1.011\%} & \textbf{-2.181\%} & & \\
 & Avg. & $1.544\pm0.008$ & $3.714\pm0.025$ & $3.554\pm0.041$ & $1.548\pm0.007$ & $3.69\pm0.024$ & $3.508\pm0.034$ & 0.259\% & \textbf{-0.646\%} & \textbf{-1.294\%} & &\\
\bottomrule
\end{tabular}}
\end{table}

We find that replacing the attention module with an MLP leads to a slight drop in forecasting performance but a significant increase in inference speed. For STAEformer, which uses vanilla transformer encoder blocks, the average drop in MAE, RMSE, and MAPE is 3.244\%, 3.331\%, and 4.981\%, with a FLOPS decrease of 65.135\% (see Table \ref{tab:staeformer_traffic}). For ASTGNN, which has an encoder-decoder structure, the removal of encoder attention blocks results in an MAE, RMSE, and MAPE drop of 0.970\%, 0.773\%, and 1.148\%, with a FLOPS reduction of 59.360\% (see Table \ref{tab:astgnn_traffic}). Interestingly, for ASTGNN, removing attention modules improves performance on the METR-LA and PEMS-BAY datasets, suggesting that stacking attention layers may lead to overfitting. From these results, we conclude that reducing over half of the computational cost without a loss of more than 2.5\% in performance demonstrates that Q, K, and V projections and attention mapping are not essential for modeling spatio-temporal data. Compared to STAEformer, ASTGNN experiences a smaller drop in parameters due to its attention-based decoder structure.

\begin{table}[]
\centering
\caption{Performance comparison between MLP-replace-Attention and AGS method using STAEformer. The average results of $\mathcal{H}=12$ are reported. The AGS method is implemented by us based on the algorithm provided. The improvement in MAE, RMSE, and MAPE of our method are highlighted in bold.}
\label{tab:ours_vs_localised}
\resizebox{1\textwidth}{!}{
\begin{tabular}{c|cccc|cccc}
\toprule
 & \multicolumn{4}{c|}{MLP-replace-Attention   vs. AGS 99\%} & \multicolumn{4}{c}{MLP-replace-Attention   vs. AGS 100\%} \\
 \midrule
 & MAE($\uparrow$) & RMSE($\uparrow$) & MAPE($\uparrow$) & FLOPS($\downarrow$) & MAE($\uparrow$) & RMSE($\uparrow$) & MAPE($\uparrow$) & FLOPS($\downarrow$) \\
 \midrule
PEMS04 & 0.598\% & \textbf{-0.129\%} & \textbf{-0.319\%} & 46.595\% & \textbf{-0.538\%} & \textbf{-0.611\%} & \textbf{-0.716\%} & 46.514\% \\
PEMS07 & 1.083\% & 1.049\% & 0.903\% & 55.582\% & \textbf{-0.107\%} & \textbf{-0.213\%} & 0.128\% & 55.539\% \\
PEMS08 & 3.071\% & 3.199\% & 3.218\% & 53.295\% & 1.003\% & 0.737\% & 0.447\% & 53.243\% \\
METR-LA & 1.168\% & 0.690\% & 1.282\% & 53.561\% & 0.932\% & 0.086\% & \textbf{-0.525\%} & 53.501\% \\
PEMS-BAY & 0.645\% & 0.000\% & \textbf{-0.279\%} & 54.285\% & 0.000\% & \textbf{-0.267\%} & \textbf{-1.108\%} & 54.204\% \\
\bottomrule
\bottomrule
\end{tabular}}
\end{table}

“We compare the MLP-replace-Attention STAEformer with the AGS pruning method proposed by \cite{duanLocalisedAdaptiveSpatialtemporal2023}. Two pruning thresholds, 99\% and 100\%, are evaluated that setting 99\% of spatial connections to 0 improves performance, while removing all connections (100\%) slightly decreases it as suggested in paper. Since the AGS pruning algorithm uses a binary mask, the FLOPS reduction is theoretically calculated, and no parameter reduction is achieved. Our results in Table \ref{tab:ours_vs_localised} confirm that the 99\% threshold improves STAEformer’s performance, while the 100\% threshold generally reduces it. In contrast, our MLP-replace-Attention approach achieves a more significant FLOPS reduction and improves performance on some datasets without pretraining or fine-tuning, indicating that both spatial and temporal attention contribute minimally to overall performance.”

\subsubsection{Long-term Time Series Forecasting} \label{sec:results_LTSF}
We observe a 1.835\% increase in MSE and a 1.317\% increase in MAE, along with an average reduction in FLOPS and parameters by 42.238\% and 9.051\%, respectively. Since the attention mechanism’s complexity scales quadratically with input size, the reduction in FLOPS also varies with input size. Replacing the attention mechanism with an MLP improves performance for longer output lengths in the Weather and ETT datasets, indicating that the attention module is redundant for modeling temporal correlations in these applications. We observe that for datasets with more variates, such as Traffic and Electricity, using MLP to approximate attention results in a more significant performance drop. This suggests that the potential for node-level interactions is more pronounced in datasets with more nodes, where attention mechanisms play a more important role in capturing spatial dependencies. PatchTST already reduces computational costs through its patching operation, which reduces the input size to 42 patches. Our results suggest that further simplifying the attention module does not significantly affect performance.

\begin{table}[]
\centering
\caption{Performance of PatchTST on multivariate datasets. Columns represent (1) the original model, (2) the MLP-replace-attention model, and (3) the improvement between them. MSE and MAE improvements are highlighted in bold.}
\label{tab:PatchTST_nontraffic}
\resizebox{1\textwidth}{!}{\begin{tabular}{cc|cc|cc|cccc}
\toprule
\multicolumn{2}{c|}{PatchTST/42} & \multicolumn{2}{c|}{Origin} & \multicolumn{2}{c|}{MLP-replace-attention} & \multicolumn{4}{c}{Improvement} \\
\midrule
&  & MSE & MAE & MSE & MAE & MSE($\uparrow$) & MAE($\uparrow$) & FLOPS($\downarrow$) & Params($\downarrow$) \\
 \midrule
\multirow{4}{*}{Weather} & 96 & $0.151\pm0.0005$ & $0.199\pm0.0003$ & $0.158\pm0.0003$ & $0.204\pm0.0002$ & 4.558\% & 2.815\% & 52.011\% & 21.468\% \\
 & 192 & $0.196\pm0.0003$ & $0.242\pm0.0004$ & $0.202\pm0.0008$ & $0.245\pm0.0005$ & 3.140\% & 1.262\% & 50.604\% & 13.761\% \\
 & 336 & $0.248\pm0.0006$ & $0.283\pm0.0008$ & $0.252\pm0.0004$ & $0.283\pm0.0003$ & 1.364\% & 0.113\% & 48.631\% & 8.943\% \\
 & 720 & $0.32\pm0.0008$ & $0.335\pm0.0006$ & $0.321\pm0.0004$ & $0.333\pm0.0003$ & 0.248\% & \textbf{-0.412\%} & 44.050\% & 4.623\% \\
\midrule
\multirow{4}{*}{Traffic} & 96 & $0.367\pm0.0003$ & $0.25\pm0.0002$ & $0.386\pm0.000$1 & $0.263\pm0.0002$ & 5.443\% & 4.893\% & 52.011\% & 21.468\% \\
 & 192 & $0.386\pm0.0006$ & $0.259\pm0.001$ & $0.401\pm0.0005$ & $0.268\pm0.001$ & 4.047\% & 3.632\% & 50.604\% & 13.761\% \\
 & 336 & $0.398\pm0.0004$ & $0.265\pm0.0005$ & $0.413\pm0.0001$ & $0.274\pm0.0001$ & 3.874\% & 3.414\% & 48.631\% & 8.943\% \\
 & 720 & $0.433\pm0.0013$ & $0.286\pm0.0022$ & $0.443\pm0.0007$ & $0.293\pm0.0012$ & 2.294\% & 2.309\% & 44.050\% & 4.623\% \\
\midrule
\multirow{4}{*}{Electricity} & 96 & $0.13\pm0.0003$ & $0.223\pm0.0001$ & $0.135\pm0.0001$ & $0.229\pm0.0002$ & 3.353\% & 2.468\% & 52.011\% & 21.468\% \\
 & 192 & $0.148\pm0.0003$ & $0.24\pm0.0005$ & $0.149\pm0.0001$ & $0.243\pm0.0001$ & 0.725\% & 0.962\% & 50.604\% & 13.761\% \\
 & 336 & $0.165\pm0.0004$ & $0.258\pm0.0004$ & $0.165\pm0.0006$ & $0.26\pm0.0008$ & 0.334\% & 0.721\% & 48.631\% & 8.943\% \\
 & 720 & $0.203\pm0.0008$ & $0.293\pm0.0009$ & $0.207\pm0.0006$ & $0.296\pm0.0004$ & 1.957\% & 1.144\% & 44.050\% & 4.623\% \\
 \midrule
\multirow{4}{*}{ILI} & 24 & $1.565\pm0.0974$ & $0.824\pm0.0237$ & $1.538\pm0.1079$ & $0.833\pm0.0272$ & \textbf{-0.339\%} & 1.192\% & 35.536\% & 9.629\% \\
 & 36 & $1.478\pm0.1269$ & $0.837\pm0.05$ & $1.555\pm0.1053$ & $0.853\pm0.0477$ & 3.534\% & 1.439\% & 35.210\% & 7.754\% \\
 & 48 & $1.782\pm0.0696$ & $0.915\pm0.0273$ & $1.833\pm0.1028$ & $0.918\pm0.0583$ & 4.016\% & 1.103\% & 34.889\% & 6.490\% \\
 & 60 & $1.522\pm0.2466$ & $0.845\pm0.0841$ & $1.766\pm0.1905$ & $0.914\pm0.0436$ & 14.900\% & 7.448\% & 34.574\% & 5.581\% \\
 \midrule
\multirow{4}{*}{ETTh1} & 96 & $0.376\pm0.0009$ & $0.4\pm0.0007$ & $0.378\pm0.0025$ & $0.402\pm0.0024$ & 0.436\% & 0.450\% & 33.859\% & 3.935\% \\
 & 192 & $0.413\pm0.0014$ & $0.421\pm0.0011$ & $0.415\pm0.0024$ & $0.422\pm0.0024$ & 0.388\% & 0.233\% & 31.623\% & 2.198\% \\
 & 336 & $0.427\pm0.0033$ & $0.433\pm0.0033$ & $0.427\pm0.002$ & $0.43\pm0.0021$ & \textbf{-0.071\%} & \textbf{-0.750\%} & 28.773\% & 1.322\% \\
 & 720 & $0.446\pm0.0067$ & $0.464\pm0.0053$ & $0.444\pm0.0045$ & $0.458\pm0.003$ & \textbf{-0.465\%} & \textbf{-1.257\%} & 23.201\% & 0.641\% \\
 \midrule
\multirow{4}{*}{ETTh2} & 96 & $0.275\pm0.0002$ & $0.336\pm0.0005$ & $0.277\pm0.0007$ & $0.338\pm0.0007$ & 0.631\% & 0.515\% & 33.859\% & 3.935\% \\
 & 192 & $0.339\pm0.0011$ & $0.378\pm0.0012$ & $0.34\pm0.0008$ & $0.381\pm0.0007$ & 0.203\% & 0.773\% & 31.623\% & 2.198\% \\
 & 336 & $0.328\pm0.0023$ & $0.381\pm0.0022$ & $0.331\pm0.0016$ & $0.387\pm0.0012$ & 0.808\% & 1.543\% & 28.773\% & 1.322\% \\
 & 720 & $0.378\pm0.0012$ & $0.421\pm0.0014$ & $0.383\pm0.0015$ & $0.425\pm0.0009$ & 1.015\% & 0.762\% & 23.201\% & 0.641\% \\
 \midrule
\multirow{4}{*}{ETTm1} & 96 & $0.29\pm0.0017$ & $0.342\pm0.0006$ & $0.292\pm0.0022$ & $0.346\pm0.0014$ & 0.988\% & 1.388\% & 52.011\% & 21.468\% \\
 & 192 & $0.334\pm0.0026$ & $0.37\pm0.001$ & $0.334\pm0.0008$ & $0.373\pm0.0006$ & 0.054\% & 0.755\% & 50.604\% & 13.761\% \\
 & 336 & $0.366\pm0.0011$ & $0.391\pm0.0008$ & $0.37\pm0.0042$ & $0.398\pm0.0041$ & 0.687\% & 1.344\% & 48.631\% & 8.943\% \\
 & 720 & $0.417\pm0.0022$ & $0.423\pm0.0014$ & $0.414\pm0.0026$ & $0.422\pm0.003$ & \textbf{-0.694\%} & \textbf{-0.157\%} & 44.050\% & 4.623\% \\
 \midrule
\multirow{4}{*}{ETTm2} & 96 & $0.165\pm0.001$ & $0.254\pm0.0004$ & $0.165\pm0.0006$ & $0.255\pm0.0005$ & \textbf{-0.045\%} & 0.550\% & 52.011\% & 21.468\% \\
 & 192 & $0.221\pm0.0009$ & $0.292\pm0.0008$ & $0.223\pm0.0004$ & $0.295\pm0.0005$ & 0.969\% & 0.718\% & 50.604\% & 13.761\% \\
 & 336 & $0.276\pm0.0014$ & $0.329\pm0.001$ & $0.279\pm0.0018$ & $0.331\pm0.0011$ & 0.793\% & 0.372\% & 48.631\% & 8.943\% \\
 & 720 & $0.364\pm0.0004$ & $0.383\pm0.0004$ & $0.363\pm0.0018$ & $0.384\pm0.0021$ & \textbf{-0.435\%} & 0.405\% & 44.050\% & 4.623\% \\
 \bottomrule
\end{tabular}}
\end{table}

Similar results are observed for the iTransformer structure, where MSE and MAE show minor increases of $1.622\%$ and $0.168\%$, but FLOPs and parameters drop significantly by $62.107\%$ and $59.228\%$, respectively. Replacing the attention mechanism in iTransformer for the ETT datasets leads to improved performance with a substantial decrease in inference cost. This is likely due to the small number of nodes (7), where node correlations can be neglected without significantly impacting performance. However, for Traffic and Electricity datasets, we observe a performance drop of over 10\% for an output length of 96, indicating that the impact of replacing attention with MLP is not positively correlated with the forecasting window length. In general, replacing the attention layer with the MLP layer maintains performance for LTSF tasks, demonstrating that attention is not essential for these tasks.

\begin{table}[]
\centering
\caption{Performance comparison between MLP-replace-Attention and DLinear using PatchTST42 and iTransformer. The improvement in MAE and RMSE of our method are highlighted in bold.}
\resizebox{0.5\textwidth}{!}{
\label{tab:ours_vs_dlinear}
\begin{tabular}{c|cc|cc}
\toprule
 & \multicolumn{2}{c|}{PatchTST42*   vs. DLinear} & \multicolumn{2}{c}{iTrans*   vs. DLinear} \\
 & MSE & MAE & MSE & MAE \\
 \midrule
Weather & \textbf{-13.017\%} & \textbf{-16.338\%} & \textbf{-1.060\%} & \textbf{-10.935\%} \\
Traffic & \textbf{-34.153\%} & \textbf{-28.332\%} & \textbf{-26.062\%} & \textbf{-23.370\%} \\
Electricity & \textbf{-23.004\%} & \textbf{-14.566\%} & \textbf{-9.013\%} & \textbf{-8.163\%} \\
ILI & \textbf{-35.998\%} & \textbf{-19.304\%} & \textbf{-15.713\%} & \textbf{-6.700\%} \\
ETTh1 & \textbf{-8.196\%} & \textbf{-4.843\%} & \textbf{-1.751\%} & \textbf{-2.531\%} \\
ETTh2 & \textbf{-35.931\%} & \textbf{-24.099\%} & \textbf{-27.568\%} & \textbf{-19.819\%} \\
ETTm1 & \textbf{-12.634\%} & \textbf{-5.394\%} & \textbf{-2.029\%} & \textbf{-1.615\%} \\
ETTm2 & \textbf{-23.713\%} & \textbf{-20.025\%} & \textbf{-15.448\%} & \textbf{-16.386\%} \\
\bottomrule
\end{tabular}}
\end{table}

We compare the MLP-replace-Attention versions of PatchTST and iTransformer with the lightweight forecaster DLinear. As shown in Table \ref{tab:ours_vs_dlinear}, the pruned PatchTST and iTransformer outperform DLinear across all datasets, averaged over four forecasting horizons. This suggests that the attention mechanism may effectively reduce to a simple identity mapping for long-term time series forecasting tasks. Since DLinear solely consists of linear layers, the FLOPS required is much lower than most of the SOTA models and thus is omitted in comparison.

\begin{table}[]
\centering
\caption{Performance of iTransformer on multivariate datasets. Columns represent (1) the original model, (2) the MLP-replace-attention model, and (3) the improvement between them. MSE and MAE improvements are highlighted in bold.}
\label{tab:iTransformer_nontraffic}
\resizebox{1\textwidth}{!}{\begin{tabular}{cc|cc|cc|cccc}
\toprule
\multicolumn{2}{c|}{iTransformer} & \multicolumn{2}{c|}{Origin} & \multicolumn{2}{c|}{MLP-replace-attention} & \multicolumn{4}{c}{Improvement} \\
\midrule
&  & MSE & MAE & MSE & MAE & MSE($\uparrow$) & MAE($\uparrow$) & FLOPS($\downarrow$) & Params($\downarrow$) \\
 \midrule
\multirow{4}{*}{Weather} & 96 & $0.176\pm0.0012$ & $0.216\pm0.002$ & $0.183\pm0.0005$ & $0.221\pm0.0005$ & 4.020\% & 2.403\% & 65.709\% & 65.171\% \\
 & 192 & $0.224\pm0.0014$ & $0.257\pm0.0009$ & $0.23\pm0.0004$ & $0.261\pm0.0005$ & 2.425\% & 1.405\% & 65.058\% & 64.514\% \\
 & 336 & $0.282\pm0.0019$ & $0.299\pm0.0014$ & $0.285\pm0.0002$ & $0.3\pm0.0001$ & 0.923\% & 0.288\% & 64.103\% & 63.553\% \\
 & 720 & $0.358\pm0.0016$ & $0.35\pm0.0011$ & $0.361\pm0.0002$ & $0.35\pm0.0004$ & 0.734\% & \textbf{-0.042\%} & 61.691\% & 61.123\% \\
 \midrule
\multirow{4}{*}{Traffic} & 96 & $0.393\pm0.0005$ & $0.268\pm0.0003$ & $0.437\pm0.0003$ & $0.282\pm0.0001$ & 11.100\% & 5.211\% & 77.802\% & 65.511\% \\
 & 192 & $0.413\pm0.0003$ & $0.277\pm0.0003$ & $0.449\pm0.0002$ & $0.287\pm0.0002$ & 8.789\% & 3.454\% & 77.419\% & 65.011\% \\
 & 336 & $0.425\pm0.0007$ & $0.283\pm0.0005$ & $0.464\pm0.0002$ & $0.293\pm0.0003$ & 9.048\% & 3.566\% & 76.853\% & 64.277\% \\
 & 720 & $0.458\pm0.0013$ & $0.301\pm0.0004$ & $0.495\pm0.0002$ & $0.312\pm0.0002$ & 8.138\% & 3.828\% & 75.381\% & 62.396\% \\
 \midrule
\multirow{4}{*}{Electricity} & 96 & $0.148\pm0.0003$ & $0.24\pm0.0003$ & $0.169\pm0.0001$ & $0.253\pm0.0001$ & 13.783\% & 5.326\% & 65.652\% & 58.549\% \\
 & 192 & $0.164\pm0.0008$ & $0.255\pm0.0008$ & $0.177\pm0.0001$ & $0.261\pm0.0001$ & 7.788\% & 2.321\% & 69.285\% & 63.001\% \\
 & 336 & $0.179\pm0.0007$ & $0.271\pm0.0006$ & $0.193\pm0.0002$ & $0.278\pm0.0001$ & 8.376\% & 2.601\% & 70.905\% & 65.043\% \\
 & 720 & $0.214\pm0.0069$ & $0.302\pm0.0047$ & $0.234\pm0.0005$ & $0.312\pm0.0004$ & 9.331\% & 3.325\% & 72.636\% & 67.334\% \\
 \midrule
\multirow{4}{*}{ILI} & 24 & $2.329\pm0.0427$ & $1.041\pm0.0156$ & $2.384\pm0.0211$ & $1.057\pm0.0049$ & 2.374\% & 1.535\% & 64.359\% & 63.886\% \\
 & 36 & $2.238\pm0.0306$ & $1.018\pm0.0096$ & $2.292\pm0.0325$ & $1.036\pm0.0093$ & 2.438\% & 1.732\% & 64.123\% & 63.648\% \\
 & 48 & $2.063\pm0.0761$ & $0.989\pm0.0287$ & $2.052\pm0.0209$ & $0.976\pm0.0056$ & \textbf{-0.557\%} & \textbf{-1.323\%} & 65.233\% & 64.996\% \\
 & 60 & $2.104\pm0.0214$ & $1.016\pm0.0111$ & $2.036\pm0.0182$ & $0.991\pm0.0047$ & \textbf{-3.247\%} & \textbf{-2.520\%} & 65.110\% & 64.873\% \\
 \midrule
\multirow{4}{*}{ETTh1} & 96 & $0.388\pm0.0015$ & $0.406\pm0.0009$ & $0.384\pm0.0003$ & $0.399\pm0.0003$ & \textbf{-1.078\%} & \textbf{-1.542\%} & 62.965\% & 62.482\% \\
 & 192 & $0.444\pm0.0009$ & $0.438\pm0.001$ & $0.436\pm0.0004$ & $0.429\pm0.0004$ & \textbf{-1.753\%} & \textbf{-1.983\%} & 61.196\% & 60.702\% \\
 & 336 & $0.488\pm0.0018$ & $0.459\pm0.0014$ & $0.475\pm0.0006$ & $0.448\pm0.0005$ & \textbf{-2.650\%} & \textbf{-2.418\%} & 62.401\% & 62.155\% \\
 & 720 & $0.518\pm0.0088$ & $0.498\pm0.0056$ & $0.493\pm0.0073$ & $0.481\pm0.0049$ & \textbf{-4.805\%} & \textbf{-3.418\%} & 58.989\% & 58.730\% \\
 \midrule
\multirow{4}{*}{ETTh2} & 96 & $0.301\pm0.0014$ & $0.351\pm0.0008$ & $0.296\pm0.0006$ & $0.348\pm0.0003$ & \textbf{-1.470\%} & \textbf{-0.791\%} & 59.782\% & 58.798\% \\
 & 192 & $0.379\pm0.0007$ & $0.399\pm0.0007$ & $0.377\pm0.0011$ & $0.397\pm0.0005$ & \textbf{-0.627\%} & \textbf{-0.384\%} & 56.733\% & 55.721\% \\
 & 336 & $0.424\pm0.0024$ & $0.432\pm0.001$ & $0.419\pm0.0003$ & $0.431\pm0.0002$ & \textbf{-1.142\%} & \textbf{-0.229\%} & 52.702\% & 51.665\% \\
 & 720 & $0.43\pm0.0053$ & $0.447\pm0.003$ & $0.426\pm0.0003$ & $0.445\pm0.0002$ & \textbf{-1.033\%} & \textbf{-0.599\%} & 44.306\% & 43.266\% \\
 \midrule
\multirow{4}{*}{ETTm1} & 96 & $0.343\pm0.0013$ & $0.377\pm0.0009$ & $0.332\pm0.0026$ & $0.367\pm0.0014$ & \textbf{-2.938\%} & \textbf{-2.526\%} & 59.782\% & 58.798\% \\
 & 192 & $0.381\pm0.0015$ & $0.395\pm0.001$ & $0.373\pm0.0003$ & $0.386\pm0.0003$ & \textbf{-2.122\%} & \textbf{-2.253\%} & 56.733\% & 55.721\% \\
 & 336 & $0.419\pm0.0013$ & $0.418\pm0.0006$ & $0.407\pm0.0003$ & $0.406\pm0.0005$ & \textbf{-2.874\%} & \textbf{-2.794\%} & 52.702\% & 51.665\% \\
 & 720 & $0.491\pm0.0025$ & $0.458\pm0.0013$ & $0.469\pm0.0003$ & $0.441\pm0.0003$ & \textbf{-4.431\%} & \textbf{-3.709\%} & 44.306\% & 43.266\% \\
 \midrule
\multirow{4}{*}{ETTm2} & 96 & $0.185\pm0.0006$ & $0.271\pm0.0012$ & $0.182\pm0.0002$ & $0.266\pm0.0003$ & \textbf{-1.863\%} & \textbf{-1.555\%} & 59.782\% & 58.798\% \\
 & 192 & $0.252\pm0.001$ & $0.313\pm0.0006$ & $0.247\pm0.0003$ & $0.308\pm0.0003$ & \textbf{-1.748\%} & \textbf{-1.481\%} & 56.733\% & 55.721\% \\
 & 336 & $0.314\pm0.001$ & $0.351\pm0.0004$ & $0.308\pm0.0003$ & $0.347\pm0.0001$ & \textbf{-1.820\%} & \textbf{-1.255\%} & 52.702\% & 51.665\% \\
 & 720 & $0.412\pm0.0012$ & $0.406\pm0.0009$ & $0.407\pm0.0009$ & $0.402\pm0.0005$ & \textbf{-1.192\%} & \textbf{-0.801\%} & 44.306\% & 43.266\% \\
 \bottomrule
\end{tabular}}
\end{table}

\subsection{Ablation Studies}
To better understand the contribution of attention mechanisms in different modules, we perform ablation studies comparing the effects of replacing spatial and temporal attention, as well as encoder and decoder attention. Additionally, we evaluate the impact of MLP components, including feedforward layers, layer normalization, and residual connections.

\subsubsection{Replace Spatial vs. Replace Temporal}
\begin{figure}[h]
     \centering
     \begin{subfigure}[b]{0.3\textwidth}
         \centering
         \includegraphics[width=\textwidth]{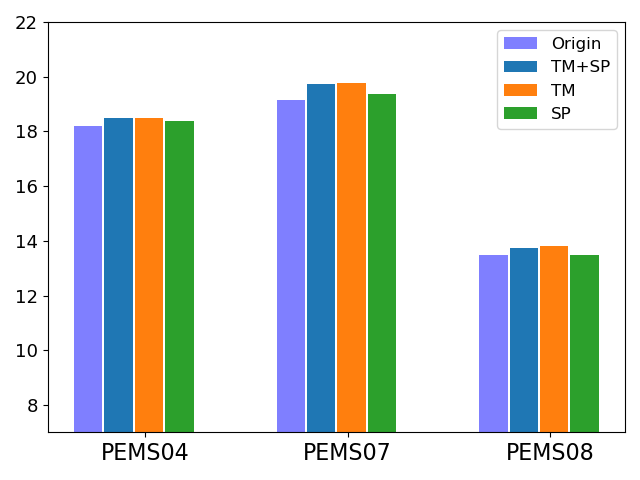}
         \caption{STAEformer MAE}
         \label{fig:staeformer_mae_pems}
     \end{subfigure}
     \hfill
     \begin{subfigure}[b]{0.3\textwidth}
         \centering
         \includegraphics[width=\textwidth]{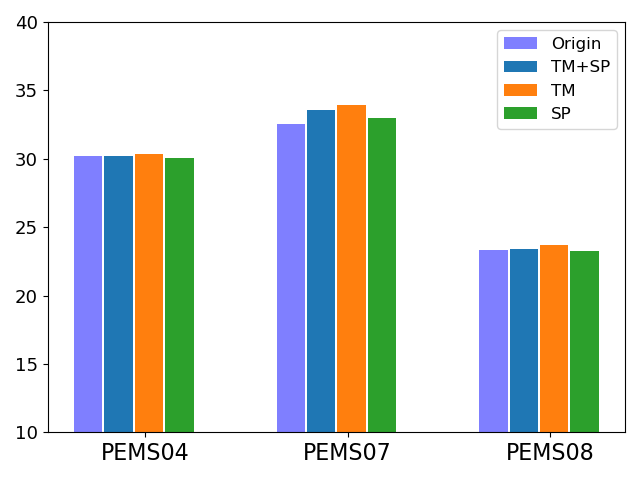}
         \caption{STAEformer RMSE}
         \label{fig:staeformer_rmse_pems}
     \end{subfigure}
     \hfill
     \begin{subfigure}[b]{0.3\textwidth}
         \centering
         \includegraphics[width=\textwidth]{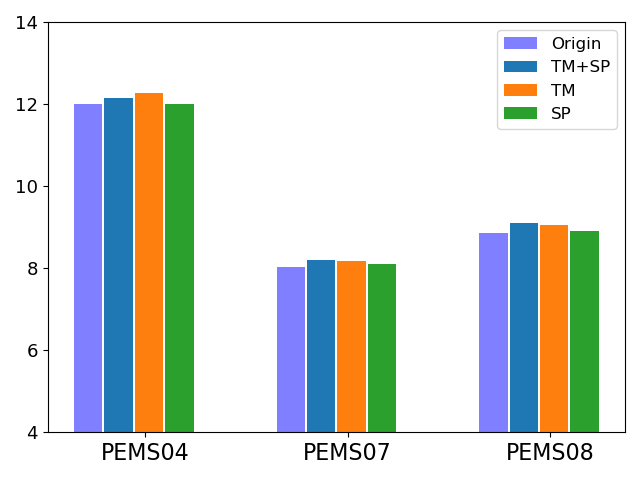}
         \caption{STAEformer MAPE}
         \label{fig:staeformer_mape_pems}
     \end{subfigure}
     \begin{subfigure}[b]{0.3\textwidth}
         \centering
         \includegraphics[width=\textwidth]{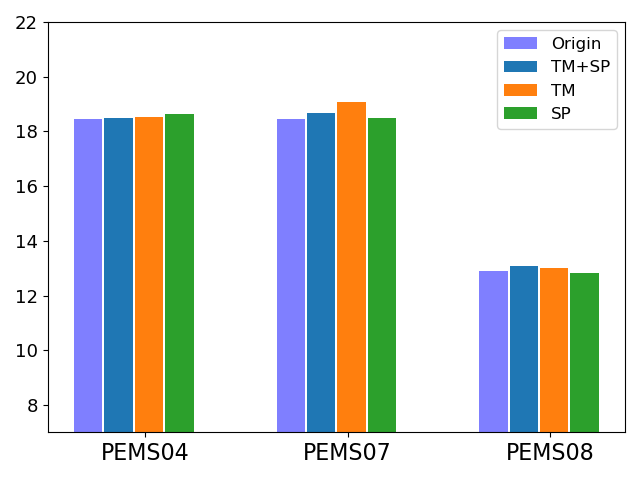}
         \caption{ASTGNN MAE}
         \label{fig:astgnn_mae_pems}
     \end{subfigure}
     \hfill
     \begin{subfigure}[b]{0.3\textwidth}
         \centering
         \includegraphics[width=\textwidth]{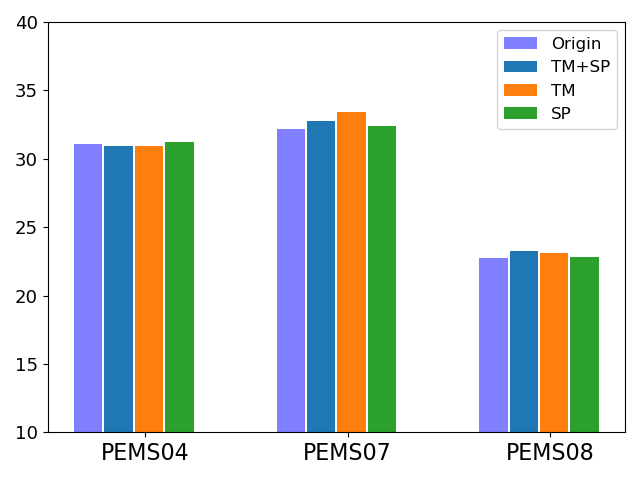}
         \caption{ASTGNN RMSE}
         \label{fig:astgnn_rmse_pems}
     \end{subfigure}
     \hfill
     \begin{subfigure}[b]{0.3\textwidth}
         \centering
         \includegraphics[width=\textwidth]{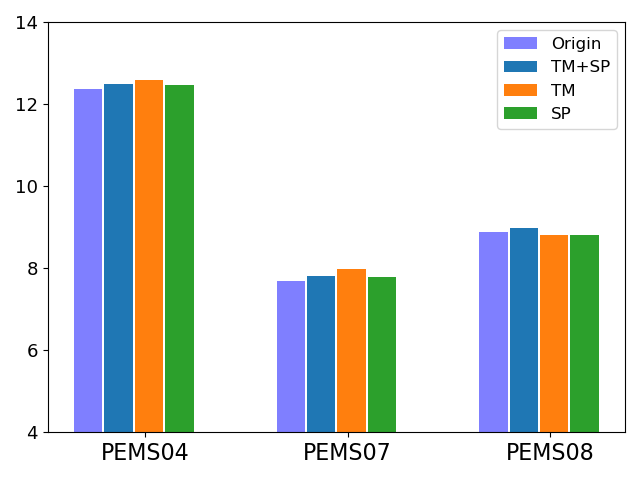}
         \caption{ASTGNN MAPE}
         \label{fig:astgnn_mape_pems}
     \end{subfigure}
     
\caption{Performance comparison of MLP layer replacement for spatial (SP) and temporal (TM) attention layers on PEMS04, 07, and 08 datasets. ‘Origin’ represents the original model, and ‘TM+SP’ denotes the model with both spatial and temporal attention layers replaced by MLP layers.}
\label{fig:spatio_vs_temporal_pems}
\end{figure}

For spatio-temporal networks using traffic datasets (PEMS04, 07, and 08), we find that removing temporal attention results in a greater performance drop than removing spatial attention. However, removing spatial attention leads to a more significant reduction in FLOPs, suggesting that spatial attention is not critical for the forecasting task in these datasets. As shown in Figure \ref{fig:spatio_vs_temporal_pems}, removing temporal attention leads to an average increase in MAE, RMSE, and MAPE of $1.904\%$, $1.876\%$, and $1.813\%$, respectively, while removing spatial attention leads to increases of $0.453\%$, $0.269\%$, and $0.417\%$. Regarding FLOPs, temporal attention contributes to a $22.441\%$ decrease, while spatial attention results in a $40.147\%$ decrease. Experiments on METR-LA and PEMS-BAY datasets can be found in the supplementary.

\subsubsection{Replace Decoder vs. Replace Encoder}
\begin{figure}
     \centering
     \begin{subfigure}[b]{0.3\textwidth}
         \centering
         \includegraphics[width=\textwidth]{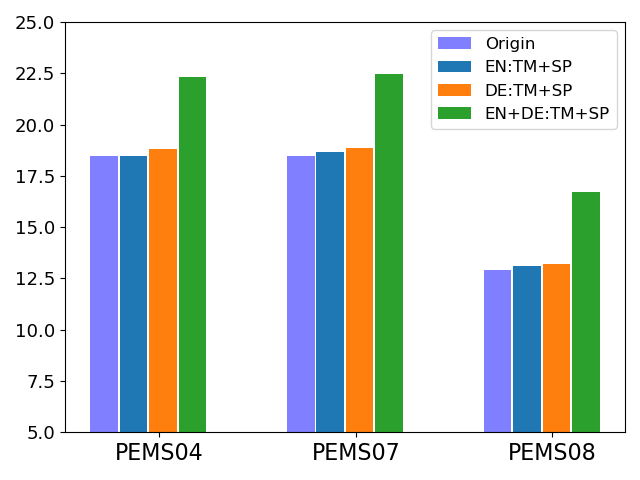}
         \caption{ASTGNN MAE}
         \label{fig:astgnn_mae_pems}
     \end{subfigure}
     \hfill
     \begin{subfigure}[b]{0.3\textwidth}
         \centering
         \includegraphics[width=\textwidth]{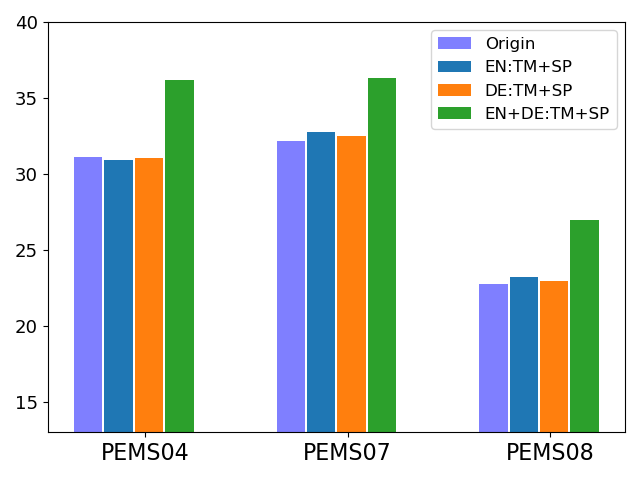}
         \caption{ASTGNN RMSE}
         \label{fig:astgnn_rmse_pems}
     \end{subfigure}
     \hfill
     \begin{subfigure}[b]{0.3\textwidth}
         \centering
         \includegraphics[width=\textwidth]{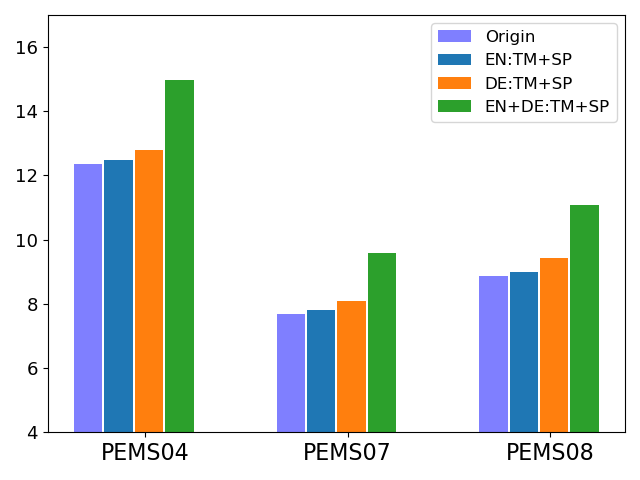}
         \caption{ASTGNN MAPE}
         \label{fig:astgnn_mape_pems}
     \end{subfigure}
\caption{Performance comparison of MLP layer replacement for encoder (EN:TM+SP) and decoder (DE:TM+SP) attention layers on PEMS04, 07, and 08 datasets. ‘Origin’ represents the original model, and ‘EN+DE:TM+SP’ denotes the model with both encoder and decoder spatial and temporal attention layers replaced by MLP layers.}
\label{fig:encoder_vs_decoder_pems}
\end{figure}

Since ASTGNN is the only model with an attention-based decoder, we conduct the attention replacement experiment on it. As shown in Figure \ref{fig:encoder_vs_decoder_pems}, replacing decoder attention results in a moderate performance drop, with a $1.944\%$, $-0.254\%$, and $3.928\%$ increase in MAE, RMSE, and MAPE, respectively. This also leads to a significant FLOPS reduction of $40.187\%$, slightly less than the $59.360\%$ FLOPS drop from replacing encoder attention. However, replacing both encoder and decoder attention simultaneously causes a substantial performance decline, with increases of $22.897\%$, $12.924\%$, and $23.313\%$ in MAE, RMSE, and MAPE. The FLOPS and parameter reductions from replacing both attention modules are $99.548\%$ and $87.886\%$, respectively, as most of the network is removed, leaving only the embedding layer as the encoder and a linear projection as the decoder. Hence, for the original encoder-decoder structure, maintaining performance requires retaining either encoder or decoder attention, but not both.

\subsubsection{FeedfForward vs. Residual vs. Layer Norm}
We compare the effects of feedforward, residual connection, and layer normalization on MLP-replace-attention models to identify the core contributors. As shown in Tables \ref{tab:ff_vs_res_vs_ln_staeformer} and \ref{tab:ff_vs_res_vs_ln_patch_itrans}, the feedforward module has the most significant impact: removing it from STAEformer results in an average increase of $29.544\%$ (MAE), $20.814\%$ (RMSE), and $48.685\%$ (MAPE). Similar results hold for LTSF tasks, where removing feedforward raises MSE and MAE by $6.168\%$ and $4.317\%$, respectively. Residual connections, while less impactful than feedforward layers, still cause performance drops—raising MAE, RMSE, and MAPE by $10.806\%$, $4.560\%$, and $37.596\%$ in STF and MSE and MAE by $2.298\%$ and $2.114\%$ in LTSF. Thus, feedforward layers and residual connections are key to MLP performance. Layer normalization, however, appears redundant. Removing it improves STAEformer performance, while PatchTST and iTransformer see minor drops under $1.2\%$. Layer normalization, primarily used to stabilize attention outputs, is less necessary without the attention layers, explaining its minimal effect on MLP models.

\begin{table}[]
\centering
\caption{Performance comparison of removing feedforward, residual connection, and layer normalization components in the MLP-replace-attention version of STAEformer on PEMS04, 07, and 08 datasets. MAE, RMSE, and MAPE improvements are highlighted in bold.}
\label{tab:ff_vs_res_vs_ln_staeformer}
\resizebox{1\textwidth}{!}{\begin{tabular}{cccccccccc}
\toprule
  & \multicolumn{3}{c}{w/o   FeedForward} & \multicolumn{3}{c}{w/o   Residual} & \multicolumn{3}{c}{w/o   LayerNorm} \\ 
  \midrule
 & MAE & RMSE & MAPE & MAE & RMSE & MAPE & MAE & RMSE & MAPE \\
 \cmidrule(lr){2-4} \cmidrule(lr){5-7} \cmidrule(lr){8-10} 
PEMS04 & 36.879\% & 29.637\% & 50.961\% & 11.442\% & 4.875\% & 46.313\% & 0.120\% & -0.017\% & \textbf{-0.007\%} \\
PEMS07 & 29.236\% & 15.788\% & 69.750\% & 17.529\% & 7.661\% & 76.133\% & \textbf{-0.077\%} & \textbf{-0.628\%} & \textbf{-0.215\%} \\
PEMS08 & 48.568\% & 34.282\% & 85.912\% & 17.672\% & 6.585\% & 59.082\% & \textbf{-0.447\%} & \textbf{-0.667\%} & \textbf{-0.159\%} \\
METR-LA & 15.125\% & 9.470\% & 14.986\% & 3.666\% & 2.142\% & 2.805\% & \textbf{-0.191\%} & \textbf{-0.303\%} & \textbf{-0.291\%} \\
PEMS-BAY & 17.912\% & 14.896\% & 21.819\% & 3.722\% & 1.539\% & 3.644\% & \textbf{-0.109\%} & \textbf{-0.285\%} & \textbf{-0.019\%} \\ 
\bottomrule
\end{tabular}}
\end{table}

\begin{table}[]
\centering
\caption{Performance comparison of removing feedforward, residual connection, and layer normalization in the MLP-replace-attention versions of PatchTST and iTransformer on multivariate datasets. Results are averaged across the four forecasting horizons. MSE and MAE improvements are highlighted in bold.}
\label{tab:ff_vs_res_vs_ln_patch_itrans}
\resizebox{0.8\textwidth}{!}{\begin{tabular}{cccccccc}
\toprule
 &  & \multicolumn{2}{c}{w/o   FeedForward} & \multicolumn{2}{c}{w/o   Residual} & \multicolumn{2}{c}{w/o   LayerNorm} \\ 
 \midrule
 &  & MSE & MAE & MSE & MAE & MSE & MAE \\ 
\cmidrule(lr){3-4} \cmidrule(lr){5-6} \cmidrule(lr){7-8} 
\multirow{8}{*}{PatchTST/42} & Weather & 7.050\% & 5.871\% & \textbf{-1.706\%} & \textbf{-1.265\%} & 0.637\% & 0.313\% \\
 & Traffic & 6.106\% & 8.377\% & 0.558\% & 1.823\% & \textbf{0.000\%} & \textbf{0.000\%} \\
 & Electricity & 3.194\% & 2.222\% & 0.321\% & 0.394\% & \textbf{-0.122\%} & 0.110\% \\
 & ILI & 12.142\% & 8.459\% & 19.813\% & 11.211\% & 16.786\% & 6.398\% \\
 & ETTh1 & \textbf{-0.718\%} & \textbf{-0.391\%} & 2.124\% & 1.755\% & 0.521\% & 0.278\% \\
 & ETTh2 & \textbf{-0.769\%} & \textbf{-0.235\%} & 0.189\% & 0.031\% & \textbf{-0.140\%} & 0.000\% \\
 & ETTm1 & 2.453\% & \textbf{-1.498\%} & 1.598\% & 0.843\% & 0.289\% & 0.199\% \\
 & ETTm2 & \textbf{-0.768\%} & \textbf{-1.146\%} & 1.259\% & 1.042\% & 0.152\% & 0.098\% \\ 
 \midrule
\multirow{8}{*}{iTransformer} & Weather & 2.127\% & 2.026\% & \textbf{-0.463\%} & \textbf{-0.680\%} & \textbf{0.000\%} & \textbf{0.000\%} \\
 & Traffic & 33.558\% & 29.555\% & 11.640\% & 15.266\% & \textbf{0.000\%} & 0.087\% \\
 & Electricity & 12.941\% & 7.081\% & 3.881\% & 3.912\% & \textbf{0.000\%} & \textbf{0.000\%} \\
 & ILI & 13.286\% & 6.984\% & \textbf{-3.074\%} & \textbf{-1.487\%} & 0.057\% & \textbf{0.000\%} \\
 & ETTh1 & 1.946\% & 0.260\% & 1.667\% & 0.926\% & \textbf{0.000\%} & \textbf{0.000\%} \\
 & ETTh2 & 2.079\% & 0.711\% & 0.305\% & 0.070\% & \textbf{0.000\%} & \textbf{0.000\%} \\
 & ETTm1 & 3.923\% & 1.060\% & \textbf{-0.149\%} & 0.741\% & \textbf{0.000\%} & \textbf{0.000\%} \\
 & ETTm2 & 0.133\% & \textbf{-0.271\%} & \textbf{-1.201\%} & \textbf{-0.752\%} & \textbf{0.000\%} & \textbf{0.000\%} \\ 
 \bottomrule
\end{tabular}}
\end{table}

\section{Conclusion}
In this work, we have shown that for spatio-temporal and long-term time series forecasting tasks, the attention module, which captures temporal and spatial patterns, can be replaced with a simple MLP that includes residual connections, layer normalization, and feedforward layers, without significantly compromising prediction performance. We introduced an abstract structure for attention-based multivariate time series forecasting models, demonstrating that the GCN modules used for spatial correlations are essentially modified attention mechanisms. Our findings suggest that the attention modules in both the encoder and decoder are not equally critical and that either can be replaced by MLP without damaging performance. Within the MLP, the feedforward layer and residual connections contribute the most to its performance.

This study has limitations. Given the wide range of attention-based models proposed for STF and LTSF tasks, we could not include all networks in our analysis, and some may still require attention mechanisms to perform optimally. Additionally, improving forecasting accuracy remains a difficult task that requires substantial effort from researchers. Our aim is not to undermine this hard work, but rather to help identify the core contributors to performance and offer new insights into model development. For future research, we suggest: 1) developing a metric to assess the cost-effectiveness of performance improvements relative to increases in FLOPS; 2) evaluating whether these findings apply to other domains such as computer vision and natural language processing; and 3) creating a metric to assess the effectiveness of attention mechanisms in large language model pruning tasks.

\section*{Acknowledgments}
This work was supported in part by the STI 2030-Major Projects of China under Grant 2021ZD0201300, and by the National Science Foundation of China under Grant 62276127.

\bibliographystyle{unsrt}  
\bibliography{MLP_replace_attention}

\end{document}